\documentclass[runningheads]{llncs}

\usepackage{eccv}

\usepackage{eccvabbrv}

\usepackage{graphicx}
\usepackage{booktabs}
\usepackage[numbers, sort&compress]{natbib}
\usepackage{hyperref}

\usepackage{amsmath,amssymb}
\usepackage{algorithm}
\usepackage{algpseudocode}
\usepackage{tcolorbox}

\usepackage{tabularx}

\newcommand{\bn}{\mathbf{n}}
\newcommand{\be}{\mathbf{e}}
\newcommand{\bP}{\mathbf{P}}
\newcommand{\bc}{\mathbf{c}}
\newcommand{\bM}{\mathbf{M}}
\newcommand{\Gcal}{\mathcal{G}}

\usepackage[utf8]{inputenc}
\usepackage[accsupp]{axessibility}  
\newtcolorbox{promptbox}{
  colback=gray!10,       
  colframe=gray!40,      
  arc=4mm,               
  boxrule=0.5pt,         
  left=10pt, right=10pt, 
  top=8pt, bottom=8pt,   
  fontupper=\ttfamily\scriptsize,   
}

\usepackage{hyperref}

\usepackage{orcidlink}

\begin{document}

\title{Physics-aware Masked Diffusion-based Flood Simulation for Urban Fisheye Disaster Detection} 

\titlerunning{PhysFlood}

\author{Sodtavilan Odonchimed\inst{1}\orcidlink{0009-0001-6823-6568} \and
Tsogt Enkhbayar\inst{1} 
\and
Oyunzul Munkhtamga\inst{2}
\and 
Munkhjargal Gochoo \inst{3}
\orcidlink{0000-0002-6613-7435}
}

\authorrunning{S.~Odonchimed et al.}

\institute{The University of Tokyo, 
\and
Mongolian University of Science and Technology,
\and 
United Arab Emirates University
}

\maketitle

\begin{abstract}
  Physical simulations that predict the behavior of urban disasters, such as climate-related flooding, play a crucial role in disaster prevention and the development of anomaly detection models. However, the severe shortage of flood data in real-world environments, combined with the inherent distortions of fisheye lens images, which are used for urban surveillance, has made high-precision simulations challenging. To address this, we propose a new physical simulation system PhysFlood that leverages Diffusion Models to synthesize realistic floods from just a single image captured by a fisheye lens. Our system not only enables simulation from a single image, but also features the ability to freely control and generate diverse flood scenarios by manipulating physically meaningful variables, such as water levels. In our evaluation experiments, we conducted a qualitative human study and demonstrated that the simulation images generated by PhysFlood exhibit both acceptable realism and robustness.
\end{abstract}

\section{Introduction}
\label{sec:intro}

Urban flooding caused by torrential rains and large-scale typhoons linked to climate change has become a frequent occurrence around the world. To mitigate these urban disasters, predicting the behavior of water and identifying potential inundation areas in advance is crucial for developing accurate hazard maps. Furthermore, disaster-related data is indispensable for training AI-based anomaly detection models used in urban infrastructure monitoring~\citep{rahnemoonfar2021floodnet, lo2021deep, kyrkou2019deep}.

However, performing such simulations in real-world environments presents two major challenges. First, there is a severe shortage of real-world flood data. Because large-scale floods fortunately do not occur frequently, gathering a sufficient amount of image data suitable for training machine learning models is extremely difficult. To address this issue, prior works such as ClimateGAN~\citep{schmidt2021climategan} and ClimateNeRF~\citep{li2023climatenerf} have attempted to generate data within simulated environments. The second challenge stems from the optical characteristics of cameras. In recent years, fisheye lens cameras with a wide field of view (FoV) have attracted attention for traffic monitoring and smart mobility applications, as a single device can monitor a vast area all at once. However, because images captured by fisheye cameras suffer from heavy geometric distortion, applying conventional image recognition techniques directly to them is difficult. Consequently, existing methods from prior research cannot be readily adapted to these scenarios.

To overcome these challenges, we propose a new physical simulation system PhysFlood that leverages Diffusion Models (DMs)~\citep{yang2023diffusion, ddpm, rombach2022high} to accommodate the specific distortions of fisheye images. Given just a single fisheye image of a normal scene, our proposed method can synthesize a realistic flood image. Furthermore, it allows users to manipulate physically meaningful variables, such as the water level, as parameters. Following the architecture of prior works, our proposed method consists of two main components: Masker and Generator. The Masker generates a 2D mask representing a physically grounded water level, while the Generator utilizes this mask and the original image to generate the final flooded image.

Through qualitative human evaluation experiments, we demonstrated that the simulation images generated by PhysFlood exhibit both acceptable realism and exceptional robustness.
\begin{figure}[t]
\begin{center}
\includegraphics[scale=0.062]{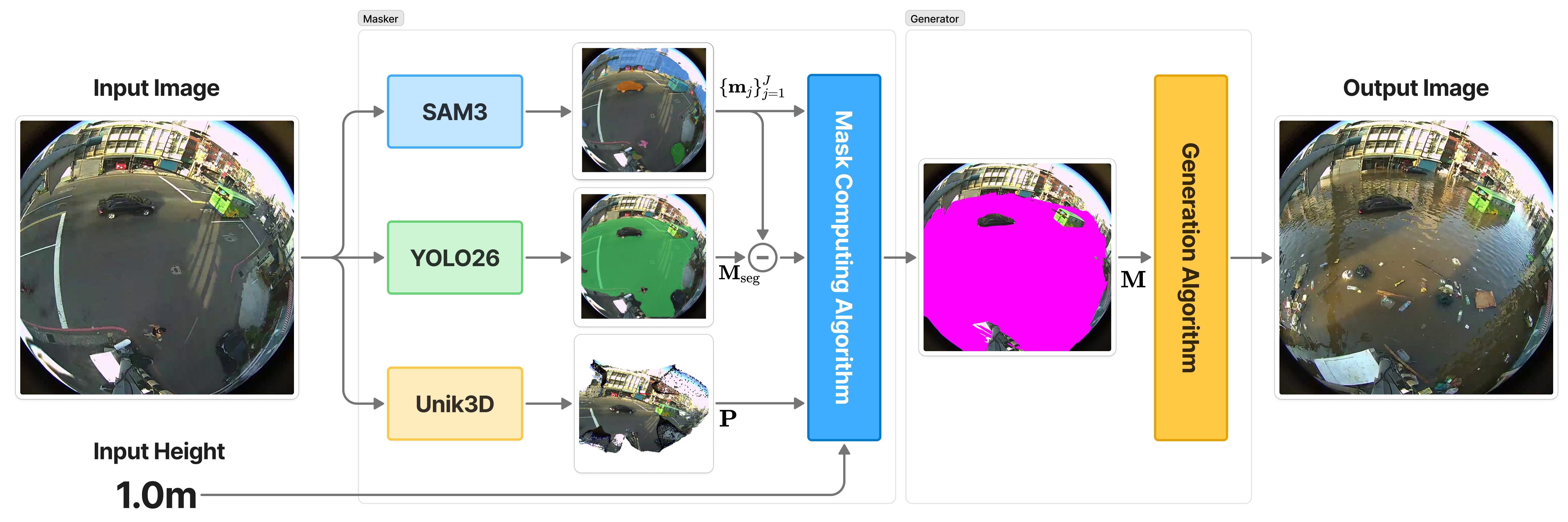}
\end{center}
\caption{\textbf{Overview of Generating Fisheye Flood Images.} PhysFlood consists of two main components: the Masker and the Generator. Within the Masker, we leverage pre-trained models, SAM3 and Unik3D, where the ground segmentation model has been explicitly trained on fisheye images. The respective outputs from these models then serve as inputs to the Mask Computing Algorithm. This algorithm also takes a user-specified flood height as input and, factoring in this constraint, computes the precise mask for the target editing region. Finally, the Generator executes the image synthesis based on the inpainting techniques commonly utilized in diffusion models.}
\label{fig:method}
\end{figure}

\section{Related Works}
\label{sec:rel}

\subsection{Image-to-Image Translation}
Image-to-Image translation has been studied since the era of GANs~\citep{goodfellow2014generative, mirza2014conditional, karras2019style, isola2017image}. These days, DMs, such as SDEdit~\citep{meng2021sdedit}, have also demonstrated their capability in this task. These methods generate a new image aligned with the original by adding appropriate noise to the base image and then denoising it. As a further advancement of SDEdit, inpainting~\citep{lugmayr2022repaint, kim2025rad} and out-painting~\citep{Song_2025_ICCV} methods have been proposed. This technique enables the editing of only specific regions within an image; by incorporating a mask during the denoising process, it restricts editing exclusively to the masked areas while preserving the unmasked regions.

\subsection{Climate Simulation}
A representative study on image generation related to climate-change is ClimateGAN, a GAN-based image-to-image translation framework~\citep{cosne2020using}. Although ClimateGAN can synthesize flood images, it cannot explicitly control the volume of the flooding. To overcome the limitations of GANs, recent studies, such as FLAME Diffuser~\citep{wang2024flame} and FloodscapeDiffuser~\citep{anwar2025floodscapediffuser}, have replaced them with DMs. These models fine-tune diffusion-based architectures on images captured from aerial or satellite perspectives, which are inherently well-suited for capturing natural disasters.
In addition to 2D image-based approaches, methods that utilize neural radiation fields (NeRF)~\citep{mildenhall2021nerf} have also been explored~\citep{xie2025climategs, qian2026weatheredit}. NeRF is a technique that reconstructs high-fidelity 3D scenes based on 2D photographs captured from multiple view angles. A prominent example of integrating NeRF into simulated disaster environments is ClimateNeRF. Using multi-view images, ClimateNeRF can render highly realistic flooded scenes while providing full controllability over the physical volume of the floodwater. In contrast, PhysFlood requires only a single image to generate results, even from a fisheye perspective.

\subsection{Fisheye Image Processing}
Fisheye images inherently suffer from geometric distortions that alter the shapes of objects, which can prevent conventional object recognition algorithms from being applied directly. For instance, in the context of object detection in fisheye imagery, several specialized methods have been developed~\citep{gochoo2023fisheye8k, Shi_2023_CVPR, kim2025edge, Rashed_2021_WACV}. Since the volume of available fisheye image datasets is severely limited compared to standard perspective datasets, some of these approaches employ DMs for data augmentation to artificially expand the training data. Alternatively, other methods propose rectifying or converting fisheye images into standard perspective images before processing~\citep{yang2021progressively}. However, because fisheye cameras possess diverse and complex intrinsic parameters, these existing rectification and augmentation techniques cannot always be readily applied to every scenario.

\section{Method}
PhysFlood is an inpainting framework that estimates physically plausible water-level masks and incorporates them into the generation process. PhysFlood consists of two main components: the Masker and the Generator. The overview is shown in ~\autoref{fig:method}.

\subsection{Masker}
The Masker takes a RGB image $I \in \mathbb{R}^{H\times W\times 3}$ and a specified flood height $H_w$ as inputs, and computes a mask $\bM \in \mathbb{R}^{H\times W}$ representing the region to be edited within the image. 

\textbf{3D Point Estimation.} To determine the heights within the image, it is necessary to estimate depth and calculate points $\bP\in\mathbb{R}^{H\times W\times 3}$ in a 3D coordinate space. And let $\bar{\bP} \in \mathbb{R}^{(HW)\times 3}$ be the reshaped points. While numerous models exist for depth estimation, converting these estimations into 3D coordinate points requires estimating the camera's intrinsic parameters. Although UniDepth ~\citep{piccinelli2024unidepth, piccinelli2025unidepthv2} is capable of predicting both depth and camera intrinsics, it has not been trained on fisheye images. To address this limitation, we employ Unik3D~\citep{piccinelli2025unik3d} which is proven effective even for fisheye imagery to perform the 3D coordinate estimation. 

\textbf{Ground Segmentation.} Next, it is necessary to predict the ground region to serve as the baseline for height. To achieve this, we utilize two separate models. Physically, when flooding occurs, water submerges the areas surrounding objects in the image. Therefore, objects within the scene must first be excluded from the ground region. We use pre-trained SAM3~\citep{carion2025sam} to generate masks $\{\mathbf{m}_j\}_{j=1} ^ J$ for objects such as cars, trucks, pedestrians, and motorcycles. We train YOLO26~\citep{sapkota2025yolo26} on a fisheye image dataset to segment only the ground areas. Finally, the precise ground region $\bM_G$ is obtained by taking the YOLO26 segmentation output $\bM_\text{seg}$ and subtracting the object masks predicted by SAM3. 
\begin{equation}
\label{alg:ground_mask}
    \bM_G = \bM_\text{seg} - \left( \bigcup_{j=1} ^J \mathbf{m}_j \right).
\end{equation}
The ground anchors are $\Gcal = \{ \bP (i,j) \mid \bM_G(i, j) = 1 \} \in \mathbb{R}^{K\times3} $.


\textbf{Height Estimation.} 
To estimate the heights, we perform a coordinate transformation denoted as $P_i =(X_i,Y_i,Z_i) \rightarrow (a_i,b_i,s_i)$. Here, $\mathbf{n}$ represents the normal vector to the ground area. Noramlized normal vector is $\mathbf{\hat{n}}=\bn / \lVert \bn \rVert$. Since the ground points $\Gcal $, are not perfectly flat, the normal vector $\mathbf{n}$ is estimated using the RANSAC algorithm~\citep{schnabel2007efficient}. Then, by selecting basis vectors $\mathbf{e}_1$ and $\mathbf{e}_2$ on the ground plane in arbitrary directions such that they satisfy $\mathbf{e}_i \cdot \mathbf{e}_j = \delta_{ij}$ and $\mathbf{e}_i \cdot \mathbf{n}=0$, the coordinate transformation can be formulated as shown in ~\autoref{eq:coord-trans}. 
\begin{equation}
    \label{eq:coord-trans}
    \bar{\bP} = a \be_1 + b \be_2 + s \hat{\textbf{n}}.
\end{equation}
Next, letting $h$ be the height estimation function. Here, $A=\left(\Gcal \cdot \be_1, \Gcal \cdot \be_2 \right) \in \mathbb{R}^{K\times 2}$ represents the ground plane, and $g = \Gcal \cdot \mathbf{\hat{n}} \in \mathbb{R}^K$ represents the height component. 
Let $Q=\left( \bar{\bP} \cdot \be_1, \bar{\bP} \cdot \be_2 \right) \in \mathbb{R}^{(HW)\times 2}$ be the ground plane at every point and let $g_\text{ref} \in \mathbb{R}^{(HW)}$ be the value obtained by interpolating $g$ on the plane $A$ with respect to all ground planes $Q$. 
If we let the interpolating function be $\mathcal{I}_{A,g}$, then $g_\text{ref}= \mathcal{I}_{A,g}(Q)$.
The function $h$ can be calculated as shown in ~\autoref{eq:height}.
\begin{equation}
    \label{eq:height}
    h(\bar{\bP}) = \bar{\bP}\cdot \mathbf{\hat{n}} - g_\text{ref}.
\end{equation}
From the ~\autoref{eq:height}, We can see that the height of the region where $\bar{\bP}$ and $\Gcal$ overlap is zero. In addition, due to the distortion of the fisheye image, $g_\text{ref}$ is not a constant value but rather a quantity that varies locally.

\begin{figure}[htbp]
\begin{center}
\includegraphics[scale=0.111]{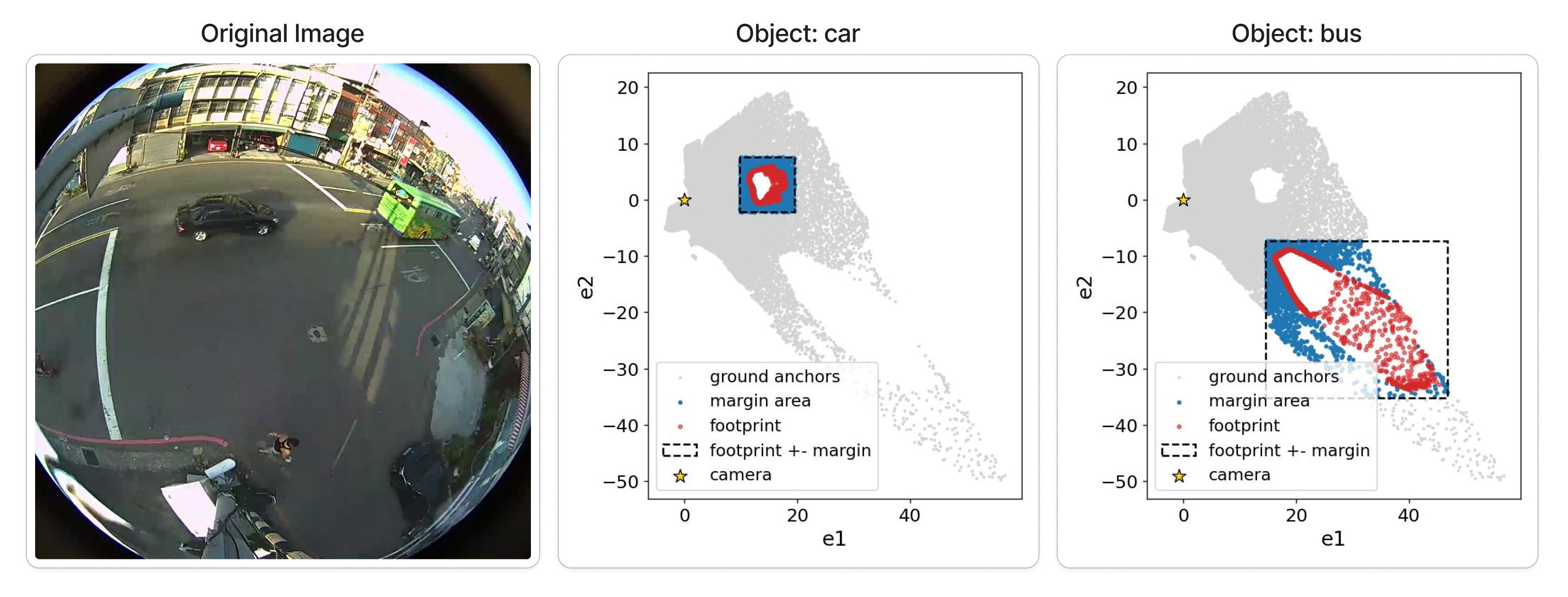}
\end{center}
\caption{\textbf{Visualization of Object Footprints and The Margin Areas.} The plotting results are shown for the black car and the green bus. Red regions represent the footprints $\mathbf{q}_j$, and blue regions denote the expanded margin areas with $\delta=1.5$.}
\label{fig:footprint}
\end{figure}

\textbf{Object Ground Interpolation.}  
Since the surface area of the ground is curved, global interpolation $g_{\text{ref}}$ can introduce errors relative to the actual object grounds. To address this, we apply revision algorithm for interpolating all object grounds. 
First, we determine the region where the object contacts the ground. For the object's point cloud $\textbf{p}_j = \{\bP(i,k)\mid \mathbf{m}_j(i,k)=1 \} \in \mathbb{R}^{N_j \times 3}$, we calculate the height components $\mathbf{s}_j = \mathbf{p}_j \cdot \hat{\mathbf{n}} \in \mathbb{R}^{N_j}$. We then extract only close to the ground points $L = \{ l \mid 1\leq l \leq N_j, \mathbf{s}_{j,l} < \tau\}$, where the points fall below a threshold $\tau$. 
We project them onto the ground plane to obtain their footprints $\textbf{q}_{j}$:
\begin{equation}
\label{eq;footprint}
  \mathbf{q}_{j,l} = (\mathbf{a}_{j,l}, \mathbf{b}_{j,l})
  = \big(\mathbf{p}_{j,l}\cdot\mathbf{e}_1,\ \mathbf{p}_{j,l}\cdot\mathbf{e}_2\big),
  \qquad l \in L,
\end{equation}
We then define the bounding box region $S$ obtained by adding a margin of $\delta$ to each side of the footprint $\textbf{q}_{j}$:
\begin{equation}
    \label{eq:footprint}
  \mathbf{u} _j
  = \Big(\min_{l\in L} \mathbf{a}_{j,l},\ \min_{l\in L} \mathbf{b}_{j,l}\Big) - \delta,
  \qquad
  \mathbf{v}_j
  = \Big(\max_{l\in L} \mathbf{a}_{j,l},\ \max_{l\in L} \mathbf{b}_{j,l}\Big) + \delta .
\end{equation}
The bounding box region $S$ is given by the ~\autoref{eq:bb_region}.
\begin{equation}
\label{eq:bb_region}
    S=\{m \mid \mathbf{u}_{j,0} \leq A_{m, 0} \leq \mathbf{v}_{j, 0}, \mathbf{u}_{j,1} \leq A_{m, 1} \leq \mathbf{v}_{j, 1} \}.
\end{equation}
~\autoref{fig:footprint} shows the results of plotting the object's footprint and the margin area.
By considering this expanded footprint margin region, it becomes possible to estimate the heights based on the local ground anchor height components that are in close proximity to the object.

\begin{figure}[htbp]
\begin{center}
\includegraphics[scale=0.18]{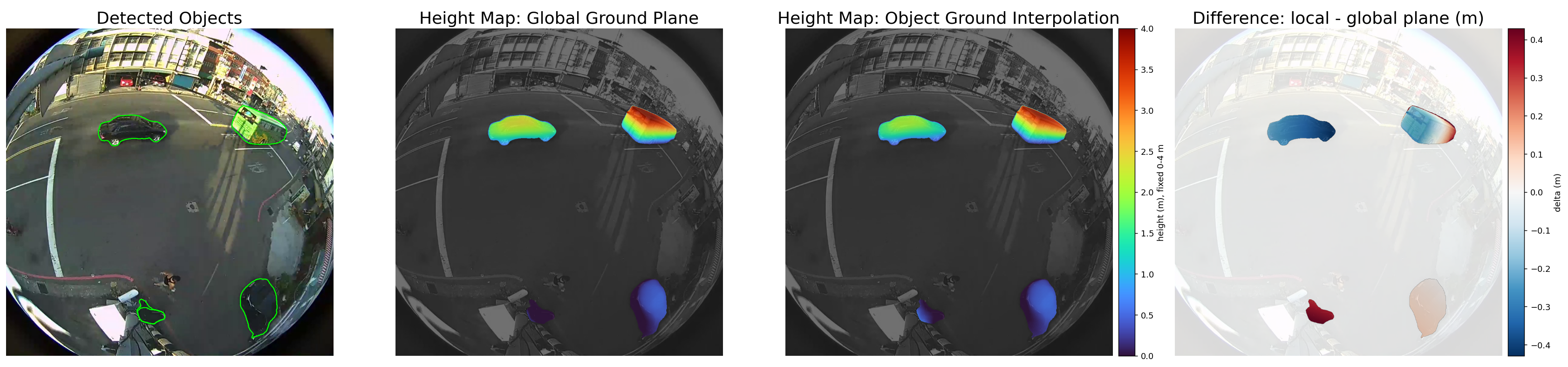}
\end{center}
\caption{\textbf{Comparison of object heights estimated from $g_{\text{ref}}$ and $g_{\text{obj}}$}. Measured object heights based on the global ground plane ($g_{\text{ref}}$) and the local ground area ($g_{\text{obj}}$) obtained via Object Ground Interpolation. Focusing on the bus, the height computed from $g_{\text{ref}}$ assumes a flat surface despite the fisheye slope. Conversely, the height computed from $g_{\text{obj}}$ accurately adapts to the local curvature of the ground.}
\label{fig:object_interpolation}
\end{figure}

Next, we retrieve the height components $g_m$ where $m\in S$.
With $g_m$ the ground level at anchor $m$, we solve
the least-squares problem
\begin{equation}
\label{eq:obj-ground-coeffs}
  \mathbf{c}^\ast
  = \arg\min_{\mathbf{c}\in\mathbb{R}^3}
    \sum_{m\in S}\Big(g_m - \big[\,A_{m,0}\ \ A_{m,1}\ \ 1\,\big]\,\mathbf{c}\Big)^2 ,
\end{equation}
yielding plane coefficients $\mathbf{c}=(c_0,c_1,c_2)$. 
By computing the coefficient $\bc$, we can determine the height component of the object's ground level.
Finally, the revised ground level is evaluated per pixel at each object point's own horizontal position, so that the reference follows the local slope across
the footprint rather than collapsing to a single offset:
\begin{equation}
\label{eq:obj-ground}
  g_\text{obj}(\mathbf{p}_j) = c_0\,(\mathbf{p}_j\cdot\mathbf{e}_1) + c_1\,(\mathbf{p}_j\cdot\mathbf{e}_2) + c_2 ,
\end{equation}
and the corrected height above ground for the object pixels is
\begin{equation}
\label{eq:obj-height}
  h(\mathbf{p}_j) = \mathbf{p}_j \cdot\hat{\mathbf{n}} - g_\text{obj}(\mathbf{p}_j).
\end{equation}
%
This per-object planar reference avoids the height estimation errors of global interpolation caused by occluded footprints on curved surfaces. As shown in \autoref{fig:object_interpolation}, using the local ground region ($g_{\text{ref}}$ $\rightarrow$ $g_{\text{obj}}$) via object ground interpolation reduces these errors compared to the global baseline ($g_{\text{ref}}$).

Once the heights are determined for all regions, the regions with a height less than or equal to the commanded height $H_w$ become the targets for editing.
\begin{equation}
    \bM = \{(i,j) \mid h(\bP(i,j)) < H_w \}.
\end{equation}

\subsection{Generator}
\label{sec:generator}
Generator edits the mask area $\bM$ computed in the Masker.

\textbf{Diffusion Models.}
Pre-trained DMs such as SD3.5~\citep{esser2024scaling} and FLUX.1~\citep{labs2025flux} are widely adopted for high-quality image generation. They generalize poorly to fisheye imagery, yet produce distorted or non-realistic results on the strong radial geometry that such lenses induce.
To close this gap, we adapt these models with Low-Rank Adaptation (LoRA)~\citep{hu2022lora}, which keeps the large pre-trained weights frozen and trains only a small set of low-rank update matrices.
This lets the network learn the geometric distortions characteristic of fisheye lenses at limited computational and memory cost, without the full-model fine-tuning that the size of these backbones would otherwise require.

\textbf{Inpainting.} 
Our Generator synthesizes flood scenes via inpainting using the LoRA-adapted DMs above.
Let $T$ be the total number of denoising steps, and let $t^*\!\le\!T$ be the step from which sampling starts, the noise level applied to the conditioning image.
Starting from $x_{t^*}$, the model denoises iteratively while constraining the unmasked region to remain faithful to the input:
\begin{equation}
\label{eq:inpainting}
x_{t-1} = \bM \odot \text{Denoise}(x_t, t) + (1 - \bM) \odot x_{t-1}^{\text{known}},
\qquad t = t^*, \dots, 1,
\end{equation}
where $\bM$ marks the region to be synthesized (the flood region) and $x_{t-1}^{\text{known}}$ is the original image $I$ diffused to noise level $t-1$:
\begin{equation}
\label{eq:inpaint}
x_{t-1}^{\text{known}} = \alpha_{t-1} I + \sigma_{t-1}\epsilon, \quad \epsilon \sim \mathcal{N}(0, \mathbf{I}).
\end{equation}
The coefficients $\alpha_t$ and $\sigma_t$ are determined by the specific type of DM employed, such as Variance-Exploding SDE (VE-SDE) or Variance-Preserving SDE (VP-SDE)~\citep{song2020score, karras2022elucidating}. The prompt used for image generation was as follows:
\begin{promptbox}
A realistic urban flood scene with water about \{cm\} cm deep (\{qualifier\}). Photorealistic flood water covering the ground and lower areas — muddy urban floodwater with reflections, gentle ripples, and partially submerged objects. Fisheye lens, circular image format, real photograph.
\end{promptbox} The target height $h$ (in cm) is substituted into \texttt{\{cm\}}, and the
\texttt{\{qualifier\}} placeholder is set according to:
\begin{equation}
\label{eq:qualifier}
\text{qualifier}(h) =
\begin{cases}
\text{"ankle-deep, shallow"}          & h < 20, \\
\text{"knee-deep, moderate"}          & 20 \le h < 70, \\
\text{"waist-deep, severe"}           & 70 \le h < 110, \\
\text{"chest-deep, very severe"}      & 110 \le h < 160, \\
\text{"extremely deep, catastrophic"} & h \ge 160.
\end{cases}
\end{equation}

\textbf{Data Augmentation via Foundation Image Models.} 
A key obstacle in this study is the lack of datasets that combine a fisheye perspective with flooded environments.
To overcome this constraint, we augment and process our data using NanoBanana Pro~\citep{nanobanana2026, zuo2025nano} and GPT-Image-2~\citep{openai2026gptimage2, zewde2026gpt}, whose strong instruction-following and generative quality let us synthesize varied yet realistic flood scenes to expand training coverage.
From a single fisheye source image, we generate augmented variants with prompts tailored to three flood levels: “ankle”, “knee”, and “waist” deep.

\textbf{Inpainting via Foundation Image Generation Models.}
Beyond the models fine-tuned in our local environment, we also synthesize images using the editing capabilities of these foundation models.
Since NanoBanana Pro and GPT-Image-2 lack a native inpainting interface, we shade the region defined by the mask $\bM$ in magenta and feed the following prompt template to the models:
\begin{promptbox}
Edit this image: replace the magenta-colored region with photorealistic flood water about \{cm\} cm deep (\{qualifier\}). The flood water should look like muddy urban floodwater with reflections, gentle ripples, and partially submerged objects appropriate for a \{cm\} cm water level. Keep everything outside the magenta region exactly as it is -- do not change the sky, buildings, vehicles above the waterline, or any other content. Preserve the original fisheye barrel distortion and circular image format. The output should look like a real photograph of a flooded scene, not an illustration or edited image.
\end{promptbox}
We also utilized a vanilla image-to-image approach that does not require a mask. For this configuration, the following prompt was employed:
\begin{promptbox}
Edit this image to show a realistic urban flood scene with water about \{cm\} cm 
deep (\{qualifier\}). Add photorealistic flood water covering the ground and lower 
areas of the scene -- muddy urban floodwater with reflections, gentle ripples, and 
partially submerged objects appropriate for a \{cm\} cm water level. 
Keep the upper parts of buildings, sky, and objects above the waterline exactly as 
they are. Preserve the original fisheye barrel distortion and circular image format. 
The output should look like a real photograph of a flooded scene, not an illustration.
\end{promptbox}

\subsection{Evaluation}
We evaluated and compared PhysFlood-based inpainting (with masks) against a standard Image-to-Image approach (without masks). 
We compare four generation configurations: the LoRA-fine-tuned SD3.5-medium, the LoRA-fine-tuned FLUX.1-dev, NanoBanana Pro, and GPT-Image-2.
\begin{table}[htbp]
\centering
\caption{Human evaluation rubric. Each generated image is scored on three axes,
each from 0 (worst) to 3 (best).}
\label{tab:rubric}
\scriptsize
\setlength{\tabcolsep}{4pt}
\renewcommand{\arraystretch}{1.1}
\begin{tabularx}{\linewidth}{@{}c X X X@{}}
\toprule
& \textbf{Q1: Visual convincingness} & \textbf{Q2: Physical plausibility} & \textbf{Q3: Usefulness for detection} \\
& \emph{Could this pass as a real photo?} & \emph{Does the water obey physics, with one consistent level?} & \emph{Good sample at the labeled severity?} \\
\midrule
\textbf{3} & Photorealistic; convincing under normal viewing. & Physically consistent; coherent water level and flood behavior. & Ideal sample; label and severity clearly match. \\
\addlinespace
\textbf{2} & Believable, but with noticeable synthetic artifacts. & Mostly plausible; one noticeable inconsistency. & Usable; minor ambiguity or artifacts. \\
\addlinespace
\textbf{1} & Clearly synthetic; multiple artifacts or distortions. & Multiple inconsistencies or implausible flood geometry. & Marginal; severity mismatch or distracting artifacts. \\
\addlinespace
\textbf{0} & Broken or unusable image. & Physically impossible; no coherent flood logic. & Unusable for training or evaluation. \\
\bottomrule
\end{tabularx}
\end{table}
To assess generation quality beyond automatic metrics, we conducted a human study. The raters independently scored images along three axes: visual convincingness (Q1), physical plausibility (Q2), and usefulness for flood
detection (Q3), each on a four-point ordinal scale, summarized in ~\autoref{tab:rubric}.
Images were presented in randomized order with model identity withheld from the raters, so that scores could not be biased by knowledge of the source model. The raters assigns scores for Q1, Q2, and Q3 based on the generated image and the commanded height.
We report the score per axis, and quantify inter-rater agreement with Krippendorff's $\alpha$~\citep{Hayes01042007}.

\section{Experiment}
\subsection{Experimental Setup}
We evaluate our method on the Fisheye8K dataset~\citep{gochoo2023fisheye8k}, a road-scene benchmark of 8{,}000 fisheye images captured by 18 traffic-surveillance cameras in urban environments.

\textbf{Training Data.}
We selected 32 source images from Fisheye8K and applied the augmentation procedure described in Section~\ref{sec:generator}, using NanoBanana Pro and GPT-Image-2 to synthesize flooded scenes at 3 water levels (ankle, knee, and waist) for each source image.
This yields a training set of 224 images.

\textbf{Model and Training.}
We fine-tune two diffusion backbones, SD3.5-medium and FLUX.1-dev, with LoRA. For both, we set the LoRA rank to 16 and train for 4{,}000 steps.

\textbf{Evaluation Data.}
For evaluation we held out one image from each of 12 viewpoints.
For each image we compute the flood mask and generate flooded scenes at five target water levels: 0.2\,m, 0.5\,m, 1.0\,m, 2.0\,m, and 4.0\,m. We generated evaluation data for 8 cases, with and without masks across 4 models, and used a total of 480 images as evaluation images for validation. 

\textbf{Generation Parameters.} 
For Object Ground Interpolation, we set the threshold $\tau$ to the 20th percentile of the object's height components, and used a margin of $\delta = 1.5$. For inpainting, we generated the images using a strength of 0.8.

\subsection{Experimental Results}
In this section, we present the experimental results of PhysFlood (with mask) and Image-to-Image (without mask).

\textbf{Inter-rater Agreement.} 
We scored Q1, Q2, and Q3 with 3 human raters. First, the calculated Krippendorff’s $\alpha$ coefficients are presented in the ~\autoref{tab:agreement}. 
\begin{table}[htbp]
\centering
\caption{Inter-rater agreement by water level (Q1: visual convincingness, Q2: physical plausibility,
Q3: usefulness for detection).}
\label{tab:agreement}
\small
\begin{tabular}{@{}lcccc@{}}
\toprule
\textbf{Water level \quad} & \textbf{Images} & \textbf{Q1} & \textbf{Q2} & \textbf{Q3} \\
\midrule
0.2\,m & 96 & 0.453 & 0.135 & 0.469 \\
0.5\,m & 96 & 0.391 & 0.101 & 0.250 \\
1.0\,m & 96 & 0.462 & 0.165 & 0.310 \\
2.0\,m & 96 & 0.459 & 0.351 & 0.409 \\
4.0\,m & 96 & 0.465 & 0.462 & 0.496 \\
\midrule
\textbf{All} & \textbf{480} & \textbf{0.445} & \textbf{0.295} & \textbf{0.397} \\
\bottomrule
\end{tabular}
\end{table}

~\autoref{tab:agreement} reports the $\alpha$ scores at each commanded water level, with the final "All" row showing the overall $\alpha$ score across the entire evaluation set. Overall, the $\alpha$ values tend to be low. This can be attributed to the fuzzy nature of the current evaluation criteria; for instance, the boundaries between scores (e.g., between 0 and 1, or between 2 and 3) are inherently subjective and ambiguous. 

Furthermore, looking at Q2 in the ~\autoref{tab:agreement}, the $\alpha$ scores for input water levels of 1.0 m or less are relatively low. This implies that poor generation quality makes it difficult and confused for raters to evaluate Q2 (Physical Plausibility). For example, objects such as vehicles serve as crucial visual references when assessing water levels between 0.2 m and 1.0 m. 
By contrast, at water levels of 2.0 m or higher, vehicles become submerged and are therefore less likely to appear in the image.
Therefore, if such objects are not cleanly synthesized in the generated images, assigning a consistent score becomes challenging.

\begin{figure}[t]
\begin{center}
\includegraphics[scale=0.23]{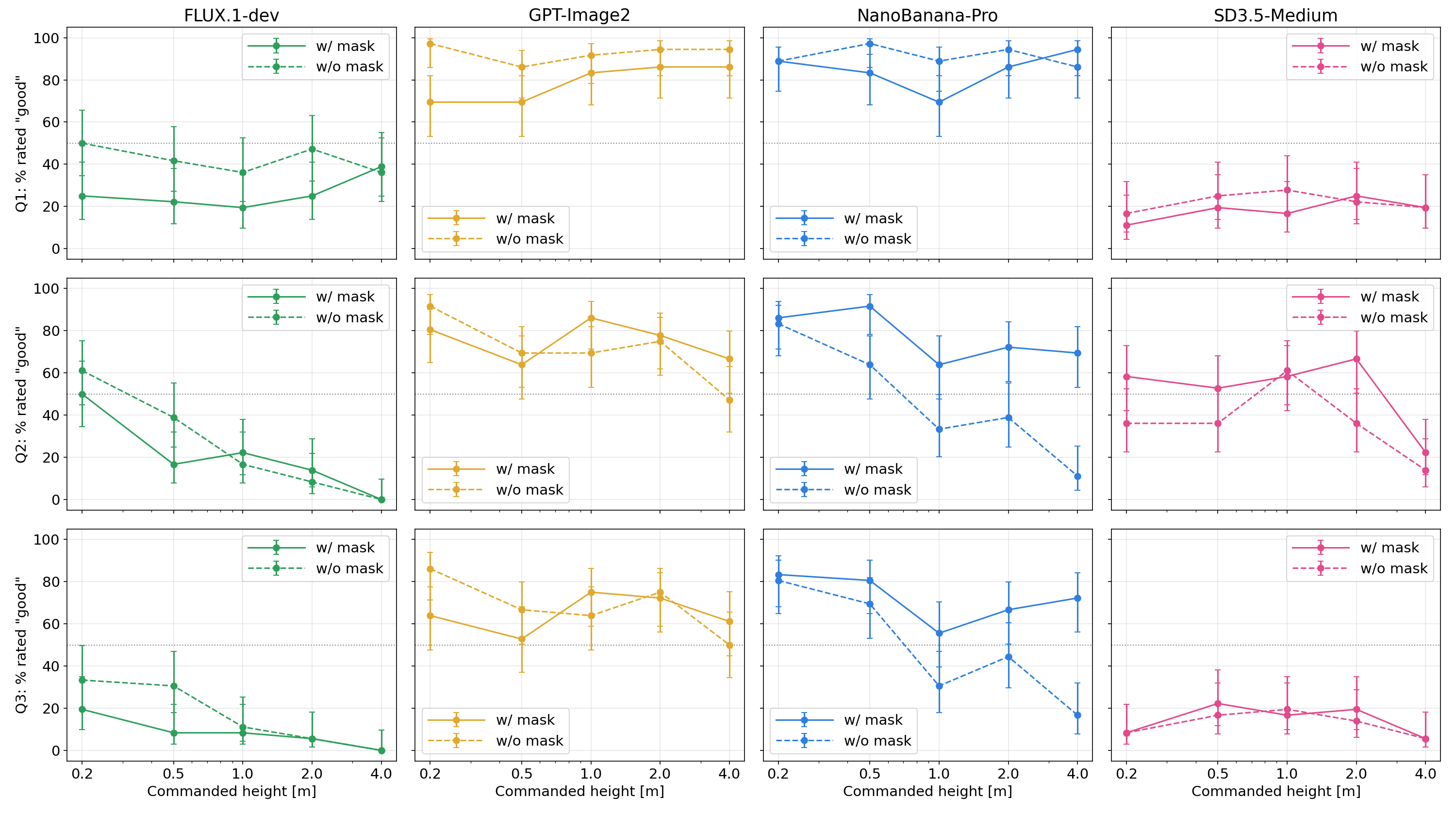}
\end{center}
\caption{\textbf{Percentage of "good" Samples.} The figure shows the scores for each model under both masked and mask-free conditions. For each question (Q1, Q2, and Q3), the scores have been relabeled into binary classes, mapping 0–1 to "bad" and 2–3 to "good". The x-axis of each plot represents the input water levels (0.2 m, 0.5 m, 1.0 m, 2.0 m, and 4.0 m), while the y-axis indicates the percentage of images classified as "good".}
\label{fig:q-scores}
\end{figure}

\textbf{Human Evaluation Result.} 
Next, we plotted the evaluation results for each model in ~\autoref{fig:q-scores}. To mitigate the impact of subjective ambiguity in the original rating scale, we relabeled the responses prior to plotting by mapping scores of 0–1 to "bad" and scores of 2–3 to "good". The figure illustrates a comparison between PhysFlood (masked) and Image-to-Image (mask-free) conditions for each model, with the y-axis representing the percentage of images classified as "good". In ~\autoref{fig:q-scores}, the two foundation models, GPT-Image-2 and NanoBanana Pro, outperform the fine-tuned models across all metrics. Focusing on Q3, GPT-Image-2 generates superior samples without a mask for water levels below 1.0 m; however, for cases with water levels of 1.0 m or higher, the masked approach yields more effective samples. In contrast, for NanoBanana Pro's Q3 results, the model with a mask successfully generates effective images across all water levels. 
\begin{figure}[t]
\begin{center}
\includegraphics[scale=0.15]{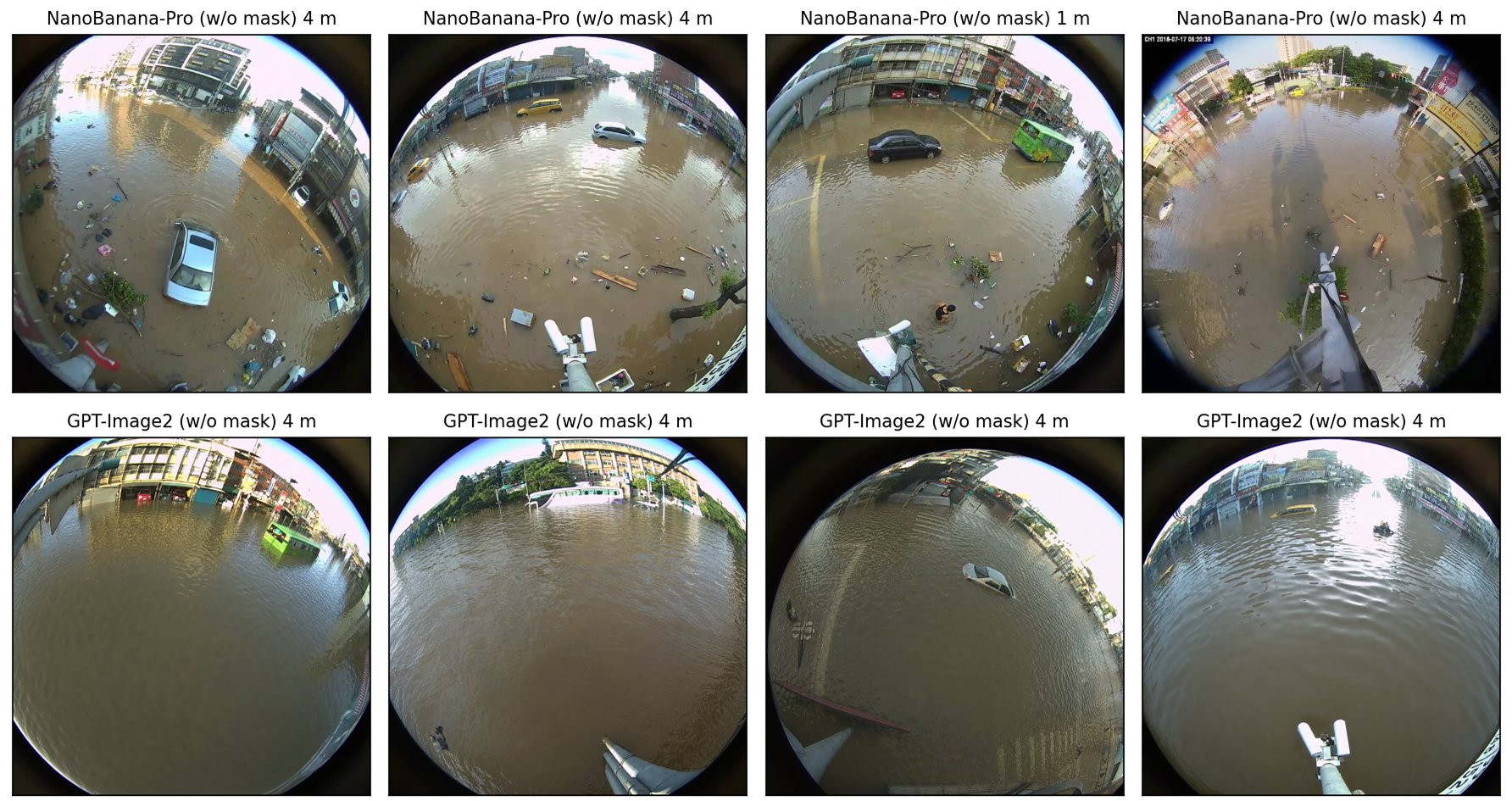}
\end{center}
\caption{\textbf{No Mask Outputs.} Example plots of the top four samples with the lowest Q2 scores for Image-to-Image generation using NanoBanana Pro and GPT-Image-2. The height indicated above each image represents the commanded water level. These examples reveal that while the generated images are realistic, the actual water levels are noticeably lower or higher than the specified input.}
\label{fig:no_masks_heights}
\end{figure}

\begin{figure}[htbp]
\begin{center}
\includegraphics[scale=0.07]{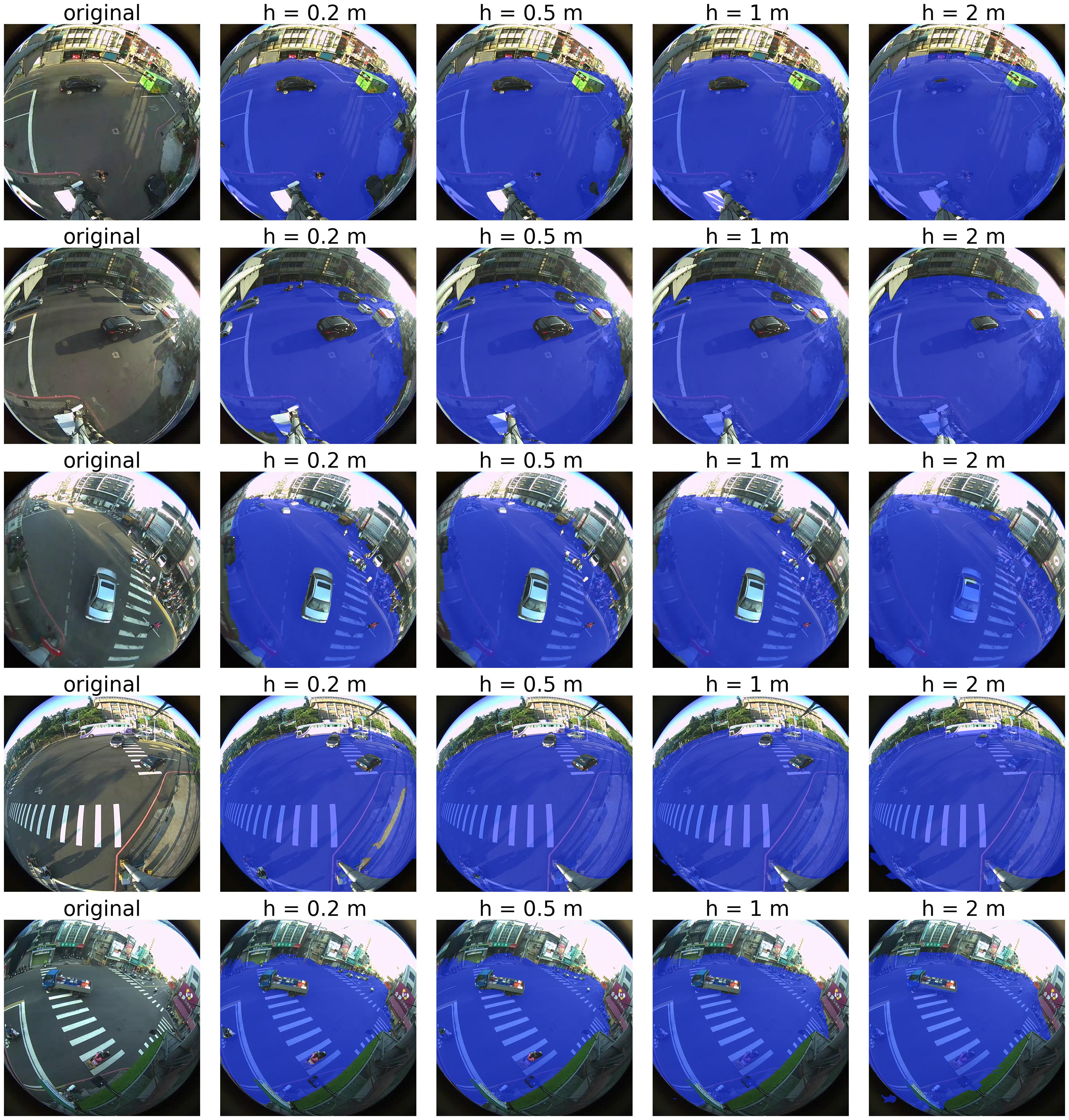}
\end{center}
\caption{\textbf{Masker Outputs.} Plotting results of the masks generated when varying the water height to 0.2m, 0.5m, 1.0m, and 2.0m. The light blue regions represent the estimated flood levels.}
\label{fig:heights}
\end{figure}

\begin{figure}[t]
\begin{center}
\includegraphics[scale=0.4]{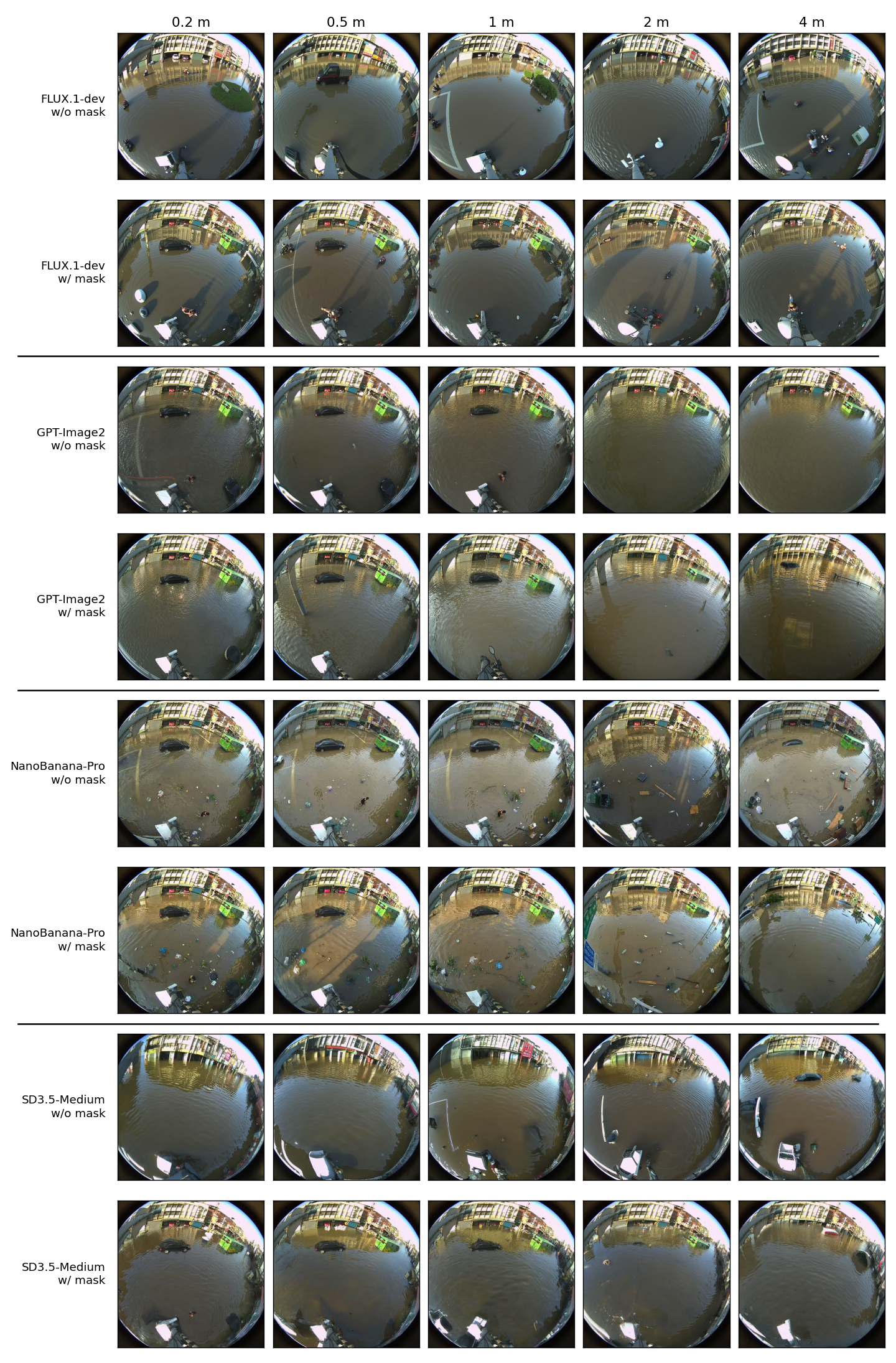}
\end{center}
\caption{\textbf{Generated Samples.} Plotting generated images under both masked and mask-free conditions for all water levels.}
\label{fig:generated-samples}
\end{figure}

\textbf{With Masks vs. Without Masks.} 
~\autoref{fig:no_masks_heights} shows that the Image-to-Image approach achieves high generation quality, but it fails to align with the commanded water levels. Furthermore, a comparison of the Q2 scores between masked and mask-free conditions for NanoBanana Pro and GPT-Image-2 in ~\autoref{fig:q-scores} shows similar results. In contrast, PhysFlood successfully generates images that match the specified commanded water levels. ~\autoref{fig:heights} and ~\autoref{fig:generated-samples} show the outputs of the Masker and Generator across various scenes.

\begin{figure}[htbp]
\begin{center}
\includegraphics[scale=0.06]{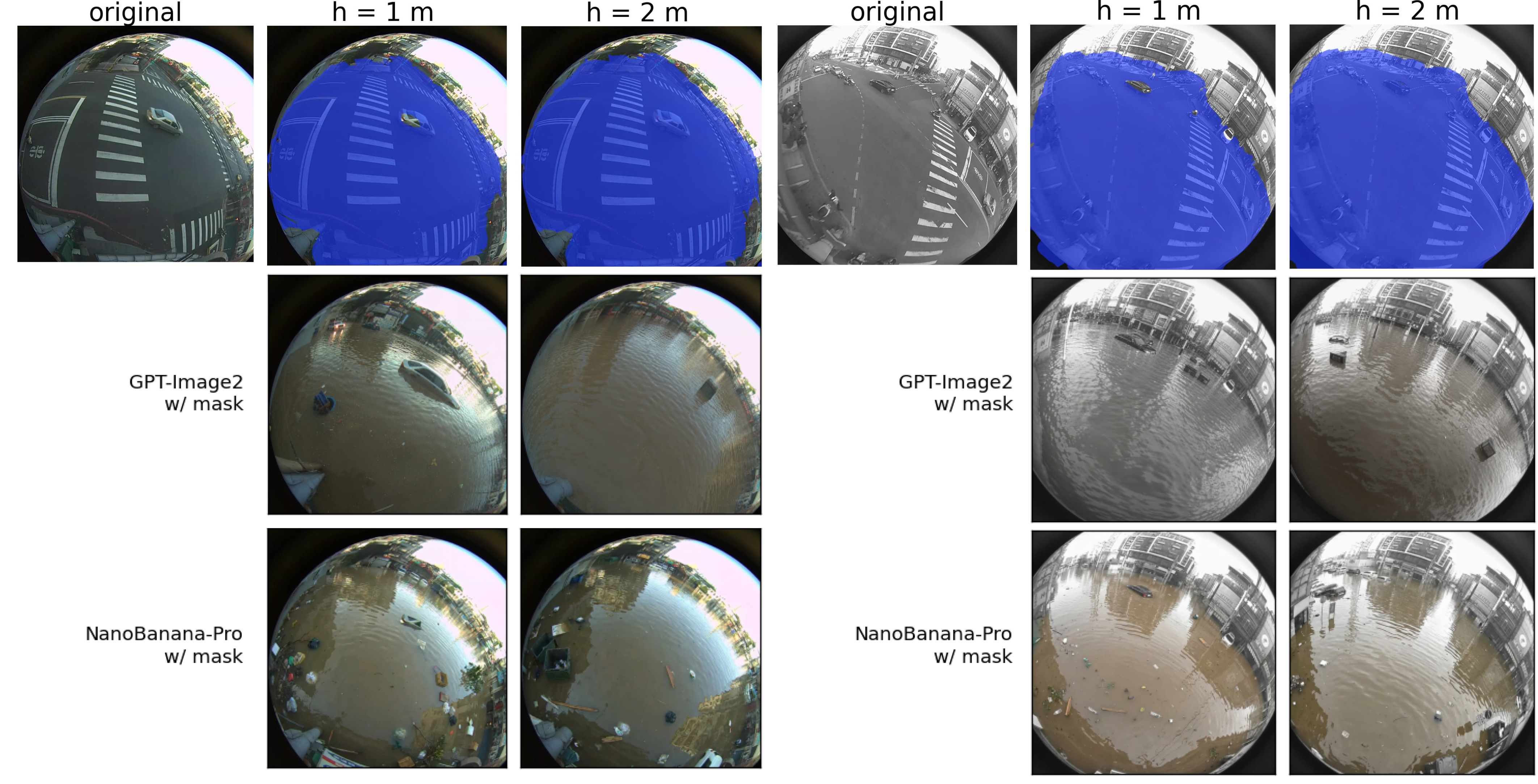}
\end{center}
\caption{\textbf{Failure Cases.} Results generated using GPT-Image-2 and NanoBanana Pro at commanded water levels of 1.0 m and 2.0 m. \textbf{Left:} Masker failure case where excessive fisheye lens tilt leads to inaccurate mask generation. \textbf{Right:} Generator failure case where the flooded regions are colored despite the grayscale input.}
\label{fig:failures}
\end{figure}

\textbf{Failure Cases.} We analyzed camera scenes where samples were not successfully generated by GPT-Image-2 and NanoBanana Pro. The following types of failure cases appeared. ~\autoref{fig:failures} displays both the masks and the generated images.
\begin{enumerate}
    \item \textbf{Masker Failure:} The rotated field of view of the fisheye camera caused the masking process to fail. Consequently, only a part of an automobile in the image was masked, leading to a visible failure in synthesizing a realistic vehicle shape in the generated image (The left side of the ~\autoref{fig:failures}).
    \item \textbf{Generator Failure:} Cases were also observed where, for gray images from night vision, only the water regions were colored (The right side of the ~\autoref{fig:failures}).
\end{enumerate}

\subsection{Limitations and Future Works}
However, this study has several limitations, which we leave as future work:
\begin{enumerate}
    \item \textbf{Number of Raters:} While three raters participated in the scoring process, engaging a larger panel of evaluators would be necessary to yield more definitive and statistically robust conclusions.
    \item \textbf{Evaluation Methodology:} Beyond human scoring, objective evaluation metrics should be explored. For assessing image quality, the adaptation of standard generative metrics such as FID (Fréchet Inception Distance) and IS (Inception Score) should be considered. For evaluating physical water levels, comparing generated heights against the ground-truth physical dimensions of reference objects presents a promising direction.
    \item \textbf{Training Strategy:} Neither of the two models fine-tuned with LoRA outperformed the foundational models. This gap is likely attributable to issues with the training data distribution. While leveraging foundation models yields superior performance, it requires higher computational costs. Future research should therefore focus on strategies for constructing high-quality training datasets specifically tailored to flooded fisheye scenarios, as well as methods to scale up both training and evaluation datasets.
\end{enumerate}

\section{Conclusion}

In this work, to address the challenge of dataset scarcity in generating flooded scenes for fisheye camera images, we proposed PhysFlood, an inpainting framework that leverages foundational image generation models and fine-tuned models trained with LoRA.

In our evaluation experiments, we conducted human subject evaluations with three raters across four configurations: SD3.5 medium, FLUX.1-dev, NanoBanana Pro, and GPT-Image-2, assessing visual persuasiveness, physical plausibility, and utility for flood detection. The experimental results confirmed that NanoBanana Pro achieved the best performance in terms of overall scores. Furthermore, while standard mask-free generation methods produced high-quality images, they failed to accurately reflect the specified commanded water levels. In contrast, PhysFlood demonstrated its ability to generate images corresponding to various water levels, even when applied to fisheye camera images.

\clearpage  


%
%
\bibliographystyle{unsrt}
\bibliography{main}
\end{document}